\documentclass{article}
     \PassOptionsToPackage{numbers, compress}{natbib}

 \usepackage[preprint]{tagds_2025}

 \workshoptitle{Topology, Algebra, and Geometry in Data Science}

\usepackage[utf8]{inputenc} 
\usepackage[T1]{fontenc}    
\usepackage{hyperref}       
\usepackage{url}            
\usepackage{booktabs}       
\usepackage{amsfonts}       
\usepackage{nicefrac}       
\usepackage{microtype}      
\usepackage{xcolor}         
\usepackage[title]{appendix}

\usepackage{amsmath,amssymb}
\usepackage{graphicx}

\usepackage{scalerel}
\usepackage{multicol}
\usepackage{multirow}

\usepackage{enumitem}
\usepackage{tabularx}
\usepackage{algorithm}
\usepackage{algorithmic}
\usepackage{csquotes}
\usepackage{array}

\usepackage{caption}

\title{Latent space analysis and generalization to out-of-distribution data}

\author{%
 Katie Rainey\thanks{Corresponding author \\ \\ Approved for public release, distribution is unlimited, AFRL-2025-0710} \\
  Naval Information Warfare Center Pacific\\
  San Diego, CA, USA \\
  \texttt{kate.e.rainey.civ@us.navy.mil} \\
  \And
  Erin Hausmann \\
  Naval Information Warfare Center Pacific \\
  San Diego, CA, USA \\
  \texttt{erin.m.hausmann.civ@us.navy.mil} \\
  \And
  Donald Waagen \\
  Air Force Research Laboratory Munitions Directorate\\
  Eglin AFB, FL, USA \\
  \And
  David Gray \\
  Air Force Research Laboratory Munitions Directorate\\
  Eglin AFB, FL, USA \\
  \And
  Donald Hulsey \\
  Dynetics, Inc. \\
  Eglin AFB, FL, USA \\  
}

\begin{document}

\maketitle

\begin{abstract}
 Understanding the relationships between data points in the latent decision space derived by the
deep learning system is critical to evaluating and interpreting the performance of the system on real
world data. Detecting \textit{out-of-distribution} (OOD) data for deep learning systems continues to be an active
research topic. We investigate the connection between latent space OOD detection and classification accuracy of the model. Using open source simulated and measured Synthetic Aperture RADAR (SAR) datasets, we empirically demonstrate that the OOD detection cannot be used as a proxy measure for model performance. We hope to inspire additional research into the geometric properties of the latent space that may yield future insights into deep learning robustness and generalizability.
\end{abstract}

\section{Introduction}

The traditional method for evaluating machine learning classification algorithms uses data which is held out from the dataset used to train the model. The idea is that the held-out data is statistically identical to the training data but not literally identical, which ensures a baseline level of generalizability. Unfortunately, this does not guarantee good performance in practical deployment. In real-world settings, it is inevitable that a model will need to make predictions on data that isn't a random split of the training data. 
However, the actual data a model sees in the real world is impossible to predict and collect before-the-fact. If we can identify when real-world data is dissimilar to the training data, we can identify when the model's performance is expected to drop. This is the basis of research into data drift and out-of-distribution (OOD) data. 

Finding these \enquote{too novel} inputs that cause poor predictions is, at its core, an anomaly detection problem. Anomaly or outlier detection has been an active topic of research in statistics and machine learning since Edgeworth \cite{Edgeworth1887}. See the surveys of Chandola \cite{Chandola2009}, Muruti  \cite{Muruti2018}, and Chalapathy and Chawla\cite{chalapathy2019}. Commercial machine learning operations monitoring tools often refer to this problem as data drift \cite{klaise2020monitoringexplainabilitymodelsproduction}. However, data drift methods only consider the data used to train the model and not the model itself. We also don't want to necessarily stop trusting the model in the presence of any data drift as that assumes the model is incapable of generalization; not all data drift impacts model performance.  

The specific problem of predicting whether a test sample comes from a different distribution from the data is often referred to as OOD detection. This differs from data drift detection in that it usually does so in the context of the classifier itself. Hendrycks \cite{hendrycks2021unsolved} coined the term OOD when he proposed a baseline method to detect OOD inputs on the basis of their confidence scores produced by the model. In a recent survey of OOD detection by Yang, et al \cite{yang2021generalized}, the authors identify two main categories of shift: semantic, where the data comes from new classes, and covariate, where the data comes from a different domain. Most algorithms and benchmarks~\cite{yang2022openood} relate only to semantic shift. 

The focus on semantic shifts in OOD detection was made explicit by \citet{ahmed2020detecting}, who argue that only semantic shifts are relevant to practical applications since classifiers should be robust to covariate shifts. But the fact that domain generalization remains an active area of research suggests that those better classifiers don’t yet exist.

Many deep learning classifiers work by encoding data into a vector space, sometimes referred to as a latent space, where different classes are able to be separated and distinguished. The hard part that requires lots of training data is finding that encoding function that maps into a space that will yield a good classifier. The training data will naturally be well classified in the latent space, but there's no guarantee that any other data will be. We therefore find the latent space an intriguing concept to study to attempt to understand how a model will classify certain data. 

In recent years, people have begun to attempt OOD detection in the latent space. \citet{Huang2020FeatureSS} make the observation that OOD samples with bounded norms will concentrate in the latent space and determine a threshold by which to detect OOD samples. \citet{muller2025mahalanobis} introduce a variation of the Mahalanobis distance used in the latent space and this variation allows for a wider variety of models. \citet{Zhang2024LearningTS} propose a novel framework that explicitly forces the underlying \textit{in-distribution} feature space to conform to a pre-defined distribution for density-based \textit{post hoc} OOD detection. Other surveys have explored using spatial relationships between test and training data in the latent space to determine if OOD such as Henze-Penrose statistic \citet{borden2022predicting} and k-nearest neighbors \citet{SunLi2022}. Overall, we feel that the use of latent space representations to understand robustness and generalizability is underexplored.

\textbf{This work was motivated by an attempt to understanding the connection between latent space geometry and classifier performance.} We leverage a dataset described by \citet{lewis2019sar} containing pairs of measured (\textit{i.e.,} real) and synthetically-generated synthetic aperture radar (SAR) images. The synthetic images were carefully constructed to match object configurations and sensor parameters from a commonly used dataset of SAR imagery. Real SAR imagery can be difficult and expensive to collect, so synthetic data is useful for augmenting training sets for classifiers. However, as in other image modalities, models trained on synthetic data do not always generalize well to real data~\citet{mumuni2024survey}, so this paired dataset was created to enable research into the synthetic-real gap for the problem of object detection in SAR imagery. In theory, a classifier trained on synthetic data should perform well on a similar real dataset, but we know that may not be true in practice. This paired dataset is ideal for testing a connection between \textit{outlierness} (via OOD detection) and classifier performance.

Starting with the synthetically-generated SAR dataset, we use an augmentation method to generate training data for a classification model.  We then consider several other datasets that the classifiers might conceivably encounter in the real world including  unaugmented synthetic SAR imagery, real SAR imagery, real SAR data at a different collection geometry than the training data, non-SAR imagery, and random noise. We use a proposed OOD detection method to identify which of these datasets contain outliers, and we determine which of these datasets should be considered \textit{in-task} based on their accuracy with the classification model. We discover that the model's ability to perform the task it has been trained for is not well correlated with the degree of \textit{outlierness} determined by the OOD algorithm.


Our paper proceeds as follows. We finish this section by discussing the motivation and preview of our findings. We detail the datasets, models, and OOD algorithm we use in Section 2. We present our experimental results and discuss the findings in Section 3. In Section 4 we conclude our paper.

\section{Experimental Design}
\label{sec:experiments} 

In this section we identify the selected model architecture and training processes, datasets, and outlier analysis tools. Additional detail can be found in Appendix \ref{app:experiments}

\subsection{Datasets}\label{sec:Datasets}

To establish the foundation for our empirical study of Deep Neural Network (DNN) outliers, we selected corresponding simulated and measured synthetic aperture radar (SAR) datasets developed by Lewis et al.~\cite{lewis2019sar} for setting up our classification problem. For the baseline classifiers of our study, we adopted the train-on-simulated, test-on-measured data approach employed by \citet{inkawhich2021bridging}, specifically for these SAR datasets. This approach relies on data augmentation techniques to address the challenge of classifier generalization to the measured data. 

Lewis's Synthetic and Measured Paired and Labeled Experiment (SAMPLE) imagery consists of a ten-class subset of the Moving and Stationary Target Acquisition and Recognition (MSTAR) measured imagery~\cite{diemunsch1998moving}, supplemented with carefully simulated versions of these MSTAR images.  For our investigation, the images were preprocessed by converting them into quarter power magnitude (QPM) representations\footnote[2]{Note that the QPM conversion was performed using code provided by Benjamin Lewis of AFRL/RY.}, cropping them to 64x64 pixel images, and mapping their pixel values onto the interval $[-1,1]$. 

\begin{figure} [h!]
   \begin{center}
   \begin{tabular}{c} 
    \includegraphics[width=5.0in]{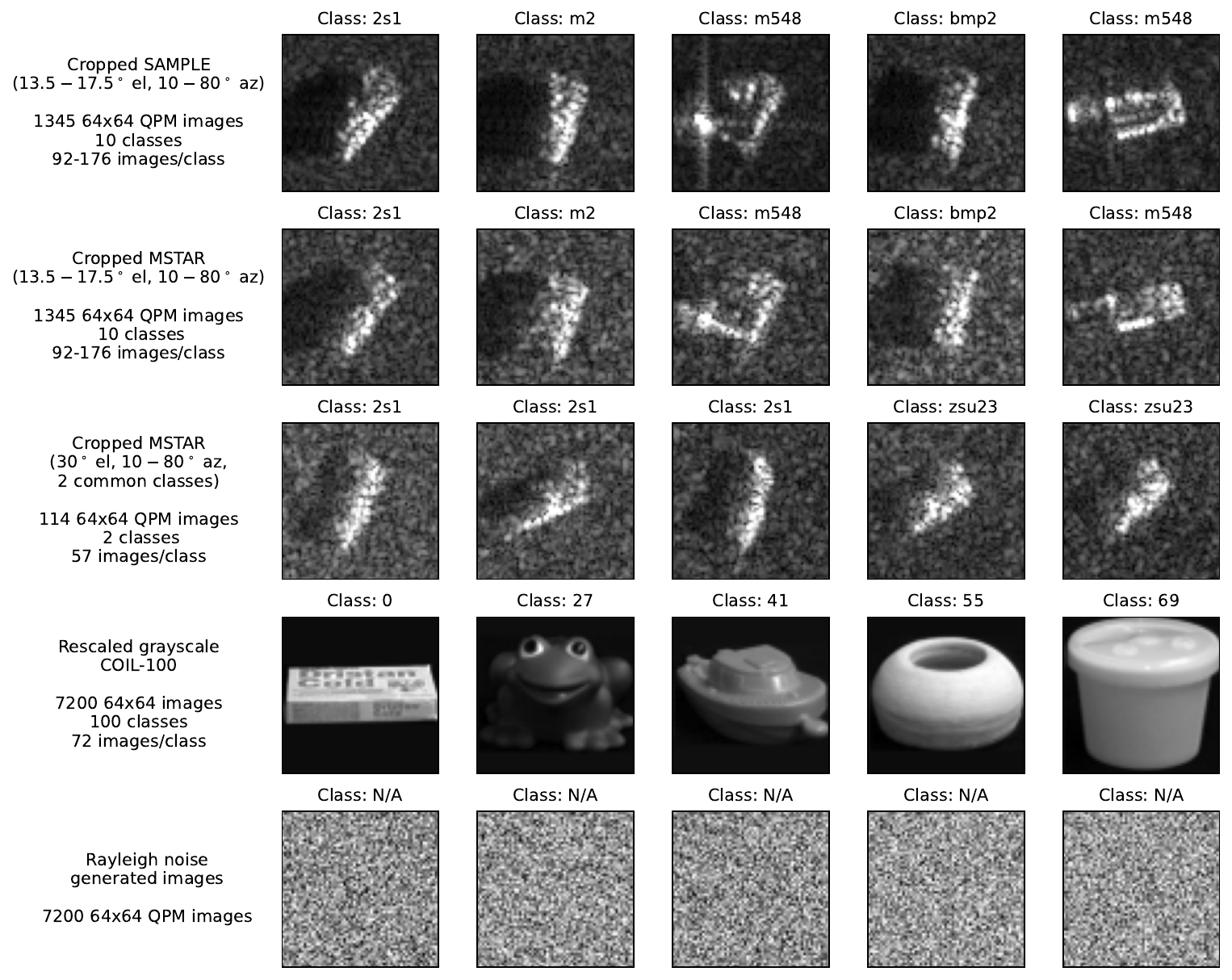}
   \end{tabular}
   \end{center}
   \caption{\label{fig:ExampleImages} Overview of Unaugmented Datasets  }
    \end{figure}

The first two rows of Figure~\ref{fig:ExampleImages} present exemplar images from these datasets, with corresponding SAMPLE simulated (top row) and MSTAR measured image (second row) pairs. As would be expected, the simulated images have more benign backgrounds than the measured. The left side of each row provides the range of azimuths (az) and elevations (el) of the targets in the image sets. As will be seen later, the choice of using the SAMPLE and MSTAR data provides insights into understanding  in-task versus OOD versus outlier situations. 

To explicitly address cases involving covariate and semantic shifts, we also included three additional datasets in our analyses. The first of these is a small two class subset of MSTAR measured SAR imagery that is limited to the same azimuth range as the previous MSTAR and SAMPLE datasets ($10-80^\circ$ az), but at a much higher elevation angle ($30^\circ$ el). Examples are shown in the third row of Figure~\ref{fig:ExampleImages}. These images undergo the same QPM preprocessing as the other SAR images. Its two classes are common to the ten-classes in the SAMPLE/MSTAR paired datasets ({\it i.e.}, no semantic shift). Among other differences, the higher elevation reduces the extent of target shadows, potentially producing covariate-shifts in the target signature distributions. 

Two additional datasets significantly deviate from the domain of our classification problem. One dataset is a modified version of the Columbia Object Image Library (COIL-100) dataset~\cite{nene96columbia}, which comprises photographs of 100 small objects at different viewing azimuths. To align them with the SAR data, the 7200 COIL-100 images were converted to grayscale and resized to 64 x 64 pixels, with pixel values mapped onto the interval [-1, 1]. The final dataset consists of 7200 64x64 pixel images of random noise pixels generated from random draws from a Rayleigh distribution, and then preprocessed into our rescaled QPM format. Exemplars from these two semantically different and phenomenology shifted datasets are shown in the fourth and fifth rows of Figure~\ref{fig:ExampleImages}.


\subsection{Models and Training}
\label{sec:ModelsAndTraining}


For our classification models, we used Keras DNN framework~\cite{keras} and the Keras software team's implementation~\cite{kerasResnet} of a  standard ResNet-20 v1 architecture~\cite{he2016deep}. Resized for this imagery, the ResNet model has a latent space with  dimensionality of 256. Note that in our usage the latent space refers to the internal output of the penultimate layer, or the last hidden layer, of a DNN classifier. For the ResNet architecture, the latent space projections are the flattened outputs of the average pooling layer. 

The training approach for the ResNet models was straightforward. Each instance of a ResNet model was trained for 500 epochs using the Adam optimizer with a learning rate of $10^{-3}$ and a batch size of 128 images. To address class size imbalance in the training, the categorical cross entropy loss was weighted inversely with dataset class size. To account for variations in initial model conditions, 50 different model instances were trained, each with distinct initial random parameter weights, for every case evaluated.

For the training data we followed the approach taken by \citet{inkawhich2021bridging}. We first divided the SAMPLE simulated imagery dataset into two elevation based subsets:\begin{itemize}
  \item  SAMPLE ($13.5^\circ - 16.5^\circ$ el, $10 - 80^\circ$ az) subset containing 806 simulated SAR images.
  \item  SAMPLE ($16.5^\circ - 17.5^\circ$ el, $10 - 80^\circ$ az) subset containing 539 simulated SAR images.
\end{itemize}
Next, we used dynamic augmentation during training of the lower elevation angle image subset ($< 16.5^\circ$ el) for input. In our image augmentation, we focused exclusively on zero-mean, additive Gaussian noise augmentation of the form:
\newline
  \begin{equation}
    \label{eq:AugmentationEq}
             x_{aug}^{(i,j)} = \text{min}\big( 1, \text{max} (-1, x_{train}^{(i,j)}+\epsilon_{i,j})\big) ,
   \end{equation}
 \newline
where $x_{train}^{(i,j)}$ and  $x_{aug}^{(i,j)}$ are the pixel $i,j$ pixel in a training image and its augmentation, respectively,  $\epsilon_{i,j} \stackrel{\text{iid}}{\sim}  \mathcal{N}(0, \sigma_{noise})$, and $ \sigma_{noise}$ is the augmentation's noise level. Note that the augmented pixel values are restricted to being on the interval [-1,1]. This augmentation approach was found to be the most significant model regularization method identified by \citet{ inkawhich2021bridging}.

\begin{figure} [ht!]
   \begin{center}
   \begin{tabular}{c} 
    \includegraphics[width=5.0in]{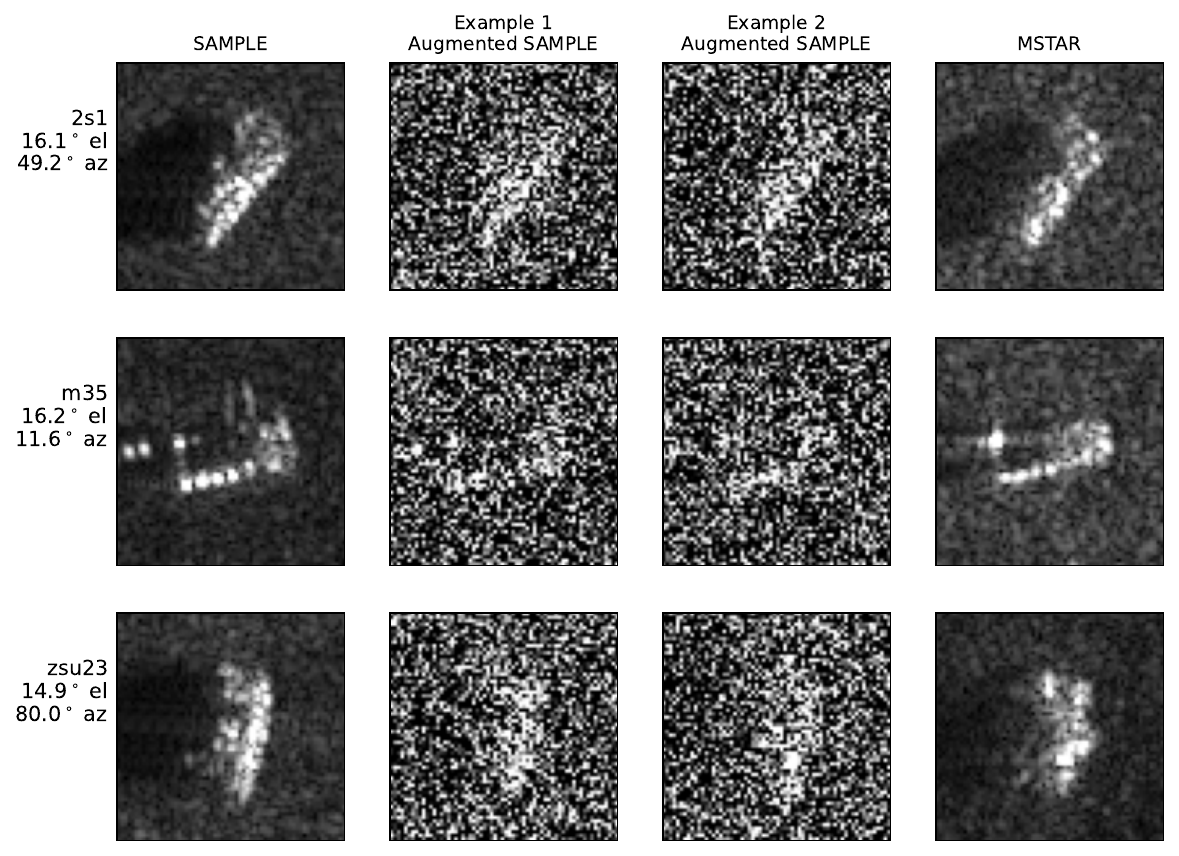}
   \end{tabular}
   \end{center}
\caption{\label{fig:ExampleAugImages2} Examples of additive Gaussian noise augmentation of SAMPLE simulated images with  augmented images with $\sigma_{noise} = 1.2$ }
 \end{figure}

In accordance with \citet{inkawhich2021bridging}, we selected the $17^\circ$ elevation  subset of the MSTAR measured imagery ($16.5 - 17.5^\circ$ el) as a  validation set for optimizing the augmentation parameter $\sigma_{noise}$. Performing a coarse search of augmentation noise level over the interval [0.0, 1.4], we discovered validation set mean class accuracies ranging from 0.11 - 0.26 for the unaugmented model instances ($\sigma_{noise} = 0$), through the highest accuracy range of 0.73 - 0.92 for $\sigma_{noise} = 1.2$  validation. This result is surprising in that it has a significantly higher preferred augmentation noise level than was identified by Inkawhich, et al., for the models that they investigated. Figure~\ref{fig:ExampleAugImages2}  shows examples of the original SAMPLE simulated SAR images (left column), their paired MSTAR images (right column), and two random augmentations of the SAMPLE images with $\sigma_{noise} = 1.2$ (the middle columns). Note the severe alteration of the images, and how different they are from not only the SAMPLE images, but also the MSTAR measured ones. Presumably, the high noise level during training teaches the classifier to be less sensitive to the background, and local pixel values.

And finally, to facilitate the analyses, we generated two supplementary static datasets of additive Gaussian noise augmented imagery ($\sigma_{noise} = 1.2$) using all images in the SAMPLE ($13.5^\circ - 16.5^\circ$ el, $10 - 80^\circ$ az) and the SAMPLE ($16.5^\circ - 17.5^\circ$ el, $10 - 80^\circ$ az) simulated data. To ensure the balance of these two sets, 1000 augmented images were generated per each of the ten classes. These are used for characterization of \enquote{outliers} as well as model classification performance.




\subsection{Deep K-Nearest Neighbors for Outlier Detection}
\label{Outlier Measures}

We used the deep $k$-nearest neighbor statistic of \citet{SunLi2022}, as a one versus many out-of-distribution outlier detection measures.  This approach demonstrated state-of-the-art performance against many competing approaches across myriad datasets.  The idea is elegant and simple; see the details in Appendix~\ref{app:knnalgo}. 
Given an encoder $\varphi$, a reference dataset $X$ of size $n$, user-selected $k$, and type 1 error threshold $\alpha$, the data $X$ are passed through the model encoder $\varphi$.  The resulting latent space representations are then normalized to unit length.  For each normalized datum in the latent space, the Euclidean distances to their respective {$k$-th nearest neighbor are computed, sorted in ascending order, and a threshold is defined by the distance corresponding to the $n*(1-\alpha)$th distance.  Any points with a $k$-th nearest neighbor distance greater than this threshold is declared {\it{out-of}} the reference distribution.

An additional measure of outlierness was considered, using the Henze-Penrose divergence to measure separation between sets of latent space representation vectors. Our results using this method were similar to those using the $k$-nearest neighbor method, so in the interest of brevity that analysis is postponed to a future publication.

%

\section{Outlier Analysis and Classification Results}\label{Analysis}


 

Let's take a moment to summarize our experimental setup and hypothesize on our results. We have trained several ResNet classifiers using augmentation on the SAMPLE dataset and tested them on the other datasets described in section \ref{sec:Datasets}. We discuss the results on models referred to as instances A, B, C, and D (see Appendix \ref{app:modelselection} for details on how these models were chosen). We also used the $k$-nearest neighbor statistic to measure whether a dataset contains outliers compared to the training dataset. The $k$-nearest neighbor statistic identifies if a dataset is out of distribution for the classifier, and classification performance indicates whether or not the dataset is out-of-task --- low performance shows that the classifier can't perform the task well. We list below the datasets we are using and our prediction of whether each is OOD or out-of-task. The subsequent sections will reveal how correct our hypotheses are.

\begin{itemize}
    \item \textbf{Augmented SAMPLE (13.5\textdegree--17.5\textdegree elevation):} Synthetic SAR imagery augmented with Gaussian noise. Ten common classes with training data. This dataset is independently drawn from the training distribution so it should be both in-distribution and in-task. 
    \item \textbf{SAMPLE (13.5\textdegree--17.5\textdegree elevation):} Synthetic SAR with no augmentation, ten common classes. Covariate shift due to the lack of augmentation. It may be identified as OOD, but should be in-task.
    \item \textbf{MSTAR (13.5\textdegree--17.5\textdegree elevation):} Measured SAR imagery, ten common classes. Covariate shift due to being measured imagery. It may be identified as OOD, but should be in-task (training augmentation was optimized for this dataset).
    \item \textbf{MSTAR (29.5\textdegree--30.5\textdegree elevation):} Measured SAR imagery, two common classes. Significant covariate shift due to being measured with a large elevation change. Expect this will be strongly identified as OOD and have a drop in classification performance (out-of-task).
    \item \textbf{COIL-100:} Greyscale images, no common classes. Semantic and covariate shifts. Expect this to be strongly OOD. Classification performance cannot be measured since there are no common classes.
    \item \textbf{Rayleigh Noise:} Noise, no common classes. Semantic and covariate shifts. Expect this to be strongly OOD. Classification performance cannot be measured since there are no common classes.
\end{itemize}


\subsection{Outlier Analysis}

To analyze the latent space using the deep $k$NN outlier detection algorithm, we selected three values for $k$ ({\it i.e.}, $k = 1, 10,\text{ and }100$) and computed the corresponding thresholds given the reference dataset (the Augmented SAMPLE dataset).  The percentages of each dataset declared as outliers (given $\alpha = 0.01$) are shown in Table \ref{tab:AllDataOutlierTable}. 

 \begin{table}[h]
\caption{Deep kNN distance outlier percentages for datasets, relative to the augmented SAMPLE dataset ($\alpha = 0.01$)} 
\label{tab:AllDataOutlierTable}
\begin{center} 
\footnotesize
\begin{tabular}{p{1in}  | c|c| p{.5in} | p{.5in} | p{.5in}|}
\cline{2-6}
   &                &   Model  & \multicolumn{3}{c|}{ Outlier percentage in set} \\
   &  Set size  & instance       & $k=1$    & $k=10$    & $k=100$    \\
\cline{1-6}
Augmented   & 10,000                      & A  &  1.6\% & 1.5\% & 1.6\% \\
\cline{3-6}
SAMPLE   & images & B  & 1.3\%  & 1.1\% & 1.0\% \\
\cline{3-6}
($16.5 - 17.5^\circ$ el,                                 & & C  & 1.6\%  & 1.5\%  &1.3\% \\
\cline{3-6}
$10 - 80^\circ$ az)                                       & & D   & 1.6\%  & 1.7\% & 1.5\% \\
\cline{2-6}
\multicolumn{6}{l}{  }  \\
\hline
SAMPLE & 1,345  &A                                              & 66.2\% & 72.4\% & 74.6\% \\
\cline{3-6}
($13.5 - 17.5^\circ$ el,  &  images & B  &  92.0\% & 94.1\% &  87.4\% \\
\cline{3-6}
$10 - 80^\circ$ az)                                       &  & C  &  48.3\% & 50.8\% &  44.6\% \\
\cline{3-6}
                                                                    &  & D  &  90.6\% & 93.9\% &  92.6\% \\
\cline{2-6}

\multicolumn{6}{l}{  }  \\
\hline
MSTAR & 1,345                                             & A   & 30.2\% & 26.6\% & 24.8\% \\
\cline{3-6}
($13.5 - 17.5^\circ$ el,  &  images & B  &  30.3\% & 28.3\% &  18.2\% \\
\cline{3-6}
 $10 - 80^\circ$ az)                                                                    &  & C  &  15.8\% & 14.7\% &   9.0\% \\
\cline{3-6}
                                                                    &  & D  &  32.1\% & 32.9\% &  26.1\% \\
 \cline{2-6}
  
\multicolumn{6}{l}{  }  \\
\hline
MSTAR                                        & 114  & A           & 87.7\% & 89.5\% & 81.6\% \\
\cline{3-6}
($29.5 - 30.5^\circ$ el,  &  images & B               &  88.6\% & 91.2\% &  71.9\% \\
\cline{3-6}
 $10 - 80^\circ$ az,                                   &  & C    &  81.6\% & 84.2\% &   64.9\% \\
\cline{3-6}
2 classes)                                                                   &  & D  &  95.6\% & 93.0\% &  82.5\% \\
 \cline{2-6}
 
 \multicolumn{6}{l}{  }  \\
\hline
Grayscale                                         & 7,200  & A  & 100.0\% & 100.0\% & 100.0\% \\
\cline{3-6}
COIL-100                                                                                  & images & B  &  100.0\% & 100.0\% &  100.0\% \\
\cline{3-6}
                                                                                  &  & C  &  100.0\% & 100.0\% &   100.0\% \\
\cline{3-6}
                                                                                  &  & D  &  100.0\% & 100.0\% &  100.0\% \\
 \cline{2-6}

 \multicolumn{6}{l}{  }  \\
\hline
Rayleigh                                                   & 7,200  & A  & 100.0\% & 100.0\% & 100.0\% \\
\cline{3-6}
noise                                                                                  &  images & B  &  100.0\% & 100.0\% &  100.0\% \\
\cline{3-6}
                                                                                  &  & C  &  100.0\% & 100.0\% &   100.0\% \\
\cline{3-6}
                                                                                  &  & D  &  100.0\% & 100.0\% &  100.0\% \\
 \cline{2-6}

\end{tabular}
\end{center}
\end{table} 

For the three values of $k$ selected, the rejection rates for the in-distribution dataset (Augmented SAMPLE) were close to the value of the error threshold $\alpha$ we selected, as should be expected.  Also as expected, the COIL-100 and Rayleigh noise datasets were marked conclusively as outliers. The other datasets had varying percentages of outliers, which will be discussed further in section \ref{sec:Conclusions}. There is significant variation in the outlier percentages  among the models, with model C rejecting at significantly lower rates for the ($13.5 - 17.5^\circ$ el, $10 - 80^\circ$ az) SAMPLE and MSTAR datasets. Additionally, for each model, the outlier rejection variation across the three values of $k$ are not considered significant, and can be explained by the reduction of the estimator variance from larger $k$ values.   Outlier percentages for all three $k$ values are shown in Table \ref{tab:AllDataOutlierTable}, but for the remainder we restrict our analysis to $k=1$.




\subsection{Classification Results}

We next look at the classification results on our various test datasets.  Classification  accuracy was computed for all fifty models.  For each SAR dataset (simulated and measured), Table~\ref{tab:AccuracySummaryTable}  in Appendix~\ref{app:modelselection} describes the minimum, mean, and maximum class accuracies achieved by the 50 model instances, but in this section we restrict our attention to the four selected model instances A, B, C, and D.   

On average, the models perform very well in correctly classifying the original SAMPLE and measured MSTAR datasets.  But how does this accuracy correlate with the an image being a distributional \enquote{inlier} or \enquote{outlier} as declared by the deep kNN?   Focusing on our four model instances (A, B, C, and D), and for each dataset under evaluation, we computed (1) the number of images declared \enquote{inlier} and \enquote{outlier}, and (2) the percentage of correct classifications given \enquote{inlier} or \enquote{outlier}.  Table~\ref{tab:K1OutlierStatsTable}
provides these results for nearest neighbor parameter $k = 1$. 

 \begin{table}[h!]
\caption{Deep kNN distance outlier statistics for correctly and incorrectly classified images ($k = 1, \alpha = 0.01$)} 
\label{tab:K1OutlierStatsTable}
\begin{center}
\footnotesize
\begin{tabular}{p{0.9in}  | c|c|c|c|c|c|c|}
\cline{2-8}
   &               &              &                 & \multicolumn{4}{c|}{ Deep kNN statistics for $k=1$ }  \\                                         
   &             &                & Images    & \multicolumn{2}{c|}{ Inlier images }  & \multicolumn{2}{c|}{ Outlier images }  \\
   &  Set      & Model     & correctly   & Total   & Correctly        & Total      & Correctly                            \\
   &  size     & instance  & classified & images   & classified    & images  & classified                            \\
\cline{1-8}
Augmented   & 10,000 &        A & 9,565 (95.7\%) & 9,838 & 95.9\% & 162 & 82.7\% \\
\cline{3-8}
SAMPLE  & images     &        B & 9,365 (93.7\%) & 9,870 & 94.0\% & 130 & 69.2\% \\
\cline{3-8}
($16.5 - 17.5^\circ$ el,  &  &   C & 9,427 (94.3\%) & 9,845 & 94.5\% & 155 & 80.0\% \\
\cline{3-8}
$10 - 80^\circ$ az)  &  &         D & 9,608 (96.1\%) & 9,836 & 96.4\% & 164 & 74.4\% \\

\cline{2-8}
\multicolumn{6}{l}{  }  \\
\hline
SAMPLE & 1,345  &              A & 1,324 (98.4\%) & 454 & 98.9\% & 891 & 98.2\% \\
\cline{3-8}
($13.5 - 17.5^\circ$ el,   &  images & B & 1224 (91.0\%) & 107 & 83.2\% & 1238 & 91.7\% \\
\cline{3-8}
$10 - 80^\circ$ az)         &  & C & 1,333 (99.1\%) & 695 & 99.9\% & 650 & 98.3\% \\
\cline{3-8}
                                       &  & D & 1,257 (93.5\%) & 127 & 96.1\% & 1,218 & 93.2\% \\
 \cline{2-8}

\multicolumn{6}{l}{  }  \\
\hline
MSTAR & 1,345                 &A & 928 (69.0\%) & 939 & 76.3\% & 406 & 52.2\% \\
\cline{3-8}
($13.5 - 17.5^\circ$ el,   &  images & B & 1074 (79.9\%) & 937 & 84.7\% & 408 & 68.6\% \\
\cline{3-8}
$10 - 80^\circ$ az)         &  & C & 1,091 (81.1\%) & 1,132 & 82.8\% & 213 & 72.3\% \\
\cline{3-8}
                                       &  & D & 1,173 (87.2\%) & 913 & 91.7\% & 432 & 77.8\% \\
 \cline{2-8}
  
\multicolumn{6}{l}{  }  \\
\hline
MSTAR & 114                     & A & 30 (26.3\%) & 14 & 35.7\% & 100 & 25.0\% \\
\cline{3-8}
($29.5 - 30.5^\circ$ el,  &    images             & B & 77 (67.5\%) & 13 & 61.5\% & 101 & 68.3\% \\
\cline{3-8}
 $10 - 80^\circ$ az,         &  & C & 51 (44.7\%) & 21 & 71.4\% & 93 & 38.7\% \\
\cline{3-8}
 2  classes)                     &  &  D & 92 (80.7\%) & 5 & 100.0\% & 109 & 79.8\% \\
 \cline{2-8}
\end{tabular}
\end{center}
\end{table}

\subsection{Discussion}

We use the results in Tables~\ref{tab:AllDataOutlierTable} and~\ref{tab:K1OutlierStatsTable} to see how our hypotheses held up.
\begin{itemize}
    \item \textbf{Augmented SAMPLE (13.5\textdegree--17.5\textdegree elevation):} As expected, this data contains few outliers and has high classification accuracy. 
    \item \textbf{SAMPLE (13.5\textdegree--17.5\textdegree elevation):} Our predictions held, as most of the images were identified as outliers, but classification accuracy remained high. 
    \item \textbf{MSTAR (13.5\textdegree--17.5\textdegree elevation):} Contains fewer outliers than SAMPLE, but has a lower classification accuracy. 
    \item \textbf{MSTAR (29.5\textdegree--30.5\textdegree elevation):} As expected, contains a high percentage of outliers and has lower classification accuracy. Interestingly, there are the most outliers with respect to model D, but that model has the highest accuracy on this dataset.
    \item \textbf{COIL-100:} As expected, all images were identified as outliers.
    \item \textbf{Rayleigh Noise:} As expected, all images were identified as outliers.
\end{itemize}

The models consistently perform well on the lower-elevation SAMPLE and MSTAR dataset, data on which the models were designed to work. These datasets are clearly \textit{in-task}. However, the OOD method frequently labels images from these sets as outliers. For the high elevation data, the OOD method is pretty convinced that the dataset is OOD, but the classification accuracy is less conclusive. The performance is degraded, but whether or not this degradation is sufficient for the data to be considered \textit{out-of-task} depends on the operational requirements of the end user. There is also notable variation in the outlier percentages depending on the model, but those variations don't map to variations in classification accuracy. Appendix~\ref{app:pvalue} considers the statistical correlation between outlier detection and classification accuracy with p-value analysis. From this data we conclude that our chosen OOD method is not a good proxy for identifying \textit{out-of-task} data.\footnote{Experiments to-be-published with another latent space based method lead to a similar conclusion.}

\section{Conclusion}
\label{sec:Conclusions}

We have implemented an OOD detection algorithm which looks for outliers among latent space vectors encoded by a deep learning classification model. The results of our investigation  indicate that data can still be \textit{in-task} with respect to a model's training, while not necessarily being \textit{in-distribution} of the training set!  We view the results as an indication that models trained for a task have the ability to generalize and handle \textit{out-of-distribution} but \textit{in-task} data.  When we observed this behavior, we endeavored to find an intrinsic geometric relationship or statistic in the latent space.  Our initial attempts to date have yet to bear fruit, but we remain steadfast that the latent space holds potential for answers.  We hope this work piques the reader's interest, and that you or a colleague can succeed in this quest.

\section{Acknowledgements}
The authors wish to extend our thanks to Benjamin Lewis of the AFRL Sensors Directorate (AFRL/RY) and Nathan Inkawhich of the AFRL Information Directorate (AFRL/RI) for their assistance in working with the SAMPLE dataset.

\bibliographystyle{plainnat} 
\bibliography{report} 

@article{Edgeworth1887,
author = {F.Y. Edgeworth},
title = {XLI. On discordant observations },
journal = {The London, Edinburgh, and Dublin Philosophical Magazine and Journal of Science},
volume = {23},
number = {143},
pages = {364--375},
year = {1887},
publisher = {Taylor \& Francis},
doi = {10.1080/14786448708628471},
URL = {https://doi.org/10.1080/14786448708628471},
eprint = {https://doi.org/10.1080/14786448708628471}
}

@inproceedings{
muller2025mahalanobis,
title={Mahalanobis++: Improving {OOD} Detection via Feature Normalization},
author={Maximilian M{\"u}ller and Matthias Hein},
booktitle={Forty-second International Conference on Machine Learning},
year={2025},
url={https://openreview.net/forum?id=vutMcZl50l}
}

@inproceedings{Zhang2024LearningTS,
  title={Learning to Shape In-distribution Feature Space for Out-of-distribution Detection},
  author={Yonggang Zhang and Jie Lu and Bo Peng and Zhen Fang and Yiu-ming Cheung},
  booktitle={Neural Information Processing Systems},
  year={2024},
  url={https://api.semanticscholar.org/CorpusID:276184995}
}

@article{Huang2020FeatureSS,
  title={Feature Space Singularity for Out-of-Distribution Detection},
  author={Haiwen Huang and Zhihan Li and Lulu Wang and Sishuo Chen and Bin Dong and Xinyu Zhou},
  journal={ArXiv},
  year={2020},
  volume={abs/2011.14654},
  url={https://api.semanticscholar.org/CorpusID:227227763}
}

@article{Chandola2009,
author = {Chandola, Varun and Banerjee, Arindam and Kumar, Vipin},
title = {Anomaly detection: A survey},
year = {2009},
issue_date = {July 2009},
publisher = {Association for Computing Machinery},
address = {New York, NY, USA},
volume = {41},
number = {3},
issn = {0360-0300},
url = {https://doi.org/10.1145/1541880.1541882},
doi = {10.1145/1541880.1541882},
journal = {ACM Comput. Surv.},
month = jul,
articleno = {15},
numpages = {58}
}

@INPROCEEDINGS{Muruti2018,
  author={Muruti, Gopinath and Rahim, Fiza Abdul and bin Ibrahim, Zul-Azri},
  booktitle={2018 IEEE Conference on Application, Information and Network Security (AINS)}, 
  title={A Survey on Anomalies Detection Techniques and Measurement Methods}, 
  year={2018},
  volume={},
  number={},
  pages={81-86},
 doi={10.1109/AINS.2018.8631436}
 }

@misc{chalapathy2019,
      title={Deep Learning for Anomaly Detection: A Survey}, 
      author={Raghavendra Chalapathy and Sanjay Chawla},
      year={2019},
      eprint={1901.03407},
      archivePrefix={arXiv},
      primaryClass={cs.LG},
      url={https://arxiv.org/abs/1901.03407}, 
}

@misc{SunLi2022,
      title={Out-of-Distribution Detection with Deep Nearest Neighbors}, 
      author={Yiyou Sun and Yifei Ming and Xiaojin Zhu and Yixuan Li},
      year={2022},
      eprint={2204.06507},
      archivePrefix={arXiv},
      primaryClass={cs.LG},
      url={https://arxiv.org/abs/2204.06507}, 
      howpublished = {\url{https://arxiv.org/abs/2204.06507}},
 }

@inproceedings{lewis2019sar,
  title={A {SAR} dataset for {ATR} development: the Synthetic and Measured Paired Labeled Experiment (SAMPLE)},
  author={Lewis, Benjamin and Scarnati, Theresa and Sudkamp, Elizabeth and Nehrbass, John and Rosencrantz, Stephen and Zelnio, Edmund},
  booktitle={Algorithms for Synthetic Aperture Radar Imagery XXVI},
  volume={10987},
  pages={39--54},
  year={2019},
  organization={SPIE}
}

@article{inkawhich2021bridging,
  title={Bridging a gap in {SAR-ATR:} Training on fully synthetic and testing on measured data},
  author={Inkawhich, Nathan and Matthew, J and Davis, Eric K and Majumder, Uttam K and Tripp, Erin and Capraro, Chris and Chen, Yiran},
  journal={IEEE Journal of Selected Topics in Applied Earth Observations and Remote Sensing},
  volume={14},
  pages={2942--2955},
  year={2021},
  publisher={IEEE}
}

@inproceedings{diemunsch1998moving,
  title={Moving and stationary target acquisition and recognition {(MSTAR)} model-based automatic target recognition: Search technology for a robust {ATR}},
  author={Diemunsch, Joseph R and Wissinger, John},
  booktitle={Algorithms for synthetic aperture radar Imagery V},
  volume={3370},
  pages={481--492},
  year={1998},
  organization={SPIE}
}

@techreport{nene96columbia,
  author = {Nayar and Murase, H.},
  institution = {Department of Computer Science, Columbia University},
  month = {February},
  number = {CUCS-006-96},
  title = {Columbia {O}bject {I}mage {L}ibrary: {COIL-100}},
  year = 1996
}

@inproceedings{he2016deep,
  title={Deep residual learning for image recognition},
  author={He, Kaiming and Zhang, Xiangyu and Ren, Shaoqing and Sun, Jian},
  booktitle={Proceedings of the IEEE conference on computer vision and pattern recognition},
  pages={770--778},
  year={2016}
}

@misc{keras,
  title={Keras},
  author={Chollet, Fran\c{c}ois and others},
  year={2015},
  howpublished={\url{https://keras.io}},
}

@misc{kerasResnet,
  author = {{Keras team}},
  title = {Trains a {ResNet} on the {CIFAR10} dataset},
  year = {2022},
  howpublished = {\url{https://keras.io/zh/examples/cifar10_resnet/}},
}

@article{hendrycks2021unsolved,
  title={Unsolved problems in ml safety},
  author={Hendrycks, Dan and Carlini, Nicholas and Schulman, John and Steinhardt, Jacob},
  journal={arXiv preprint arXiv:2109.13916},
  year={2021}
}

@inproceedings{ahmed2020detecting,
  title={Detecting semantic anomalies},
  author={Ahmed, Faruk and Courville, Aaron},
  booktitle={Proceedings of the AAAI Conference on Artificial Intelligence},
  volume={34},
  pages={3154--3162},
  year={2020}
}

@article{yang2021generalized,
  title={Generalized out-of-distribution detection: A survey},
  author={Yang, Jingkang and Zhou, Kaiyang and Li, Yixuan and Liu, Ziwei},
  journal={arXiv preprint arXiv:2110.11334},
  year={2021}
}

@article{yang2022openood,
  title={{OpenOOD}: Benchmarking generalized out-of-distribution detection},
  author={Yang, Jingkang and Wang, Pengyun and Zou, Dejian and Zhou, Zitang and Ding, Kunyuan and Peng, Wenxuan and Wang, Haoqi and Chen, Guangyao and Li, Bo and Sun, Yiyou and others},
  journal={Advances in Neural Information Processing Systems},
  volume={35},
  pages={32598--32611},
  year={2022}
}

@misc{klaise2020monitoringexplainabilitymodelsproduction,
      title={Monitoring and explainability of models in production}, 
      author={Janis Klaise and Arnaud Van Looveren and Clive Cox and Giovanni Vacanti and Alexandru Coca},
      year={2020},
      eprint={2007.06299},
      archivePrefix={arXiv},
      primaryClass={stat.ML},
      url={https://arxiv.org/abs/2007.06299}, 
}

@article{mumuni2024survey,
  title={A survey of synthetic data augmentation methods in machine vision},
  author={Mumuni, Alhassan and Mumuni, Fuseini and Gerrar, Nana Kobina},
  journal={Machine Intelligence Research},
  volume={21},
  number={5},
  pages={831--869},
  year={2024},
  publisher={Springer}
}

@inproceedings{borden2022predicting,
  title={Predicting classifier performance using distributional separation measures},
  author={Borden, Samuel T and Rainey, Katie},
  booktitle={Artificial Intelligence and Machine Learning for Multi-Domain Operations Applications IV},
  volume={12113},
  pages={343--351},
  year={2022},
  organization={SPIE}
}

\begin{appendices}
\section{Deep $k$-Nearest Neighbor Algorithm}
\label{app:knnalgo}

The details of the $k$-nearest neighbor outlier detection algorithm, taken from \citet{SunLi2022}, are shown in Algorithm~\ref{alg:Sun}.

\begin{algorithm}
   \caption{Deep $k$-Nearest Neighbors for OOD detection \citet{SunLi2022}}
   \label{alg:Sun}
   \begin{algorithmic}
   \STATE \underline{Threshold calculation:}
      \STATE Given user specified $k$, encoder $\varphi$,  type-1 error threshold parameter $\alpha$, and  reference set $\mathbf{X} = [ \mathbf{x}_1, \mathbf{x}_2, \dots, \mathbf{x}_n] $
      \STATE Compute $ \mathbf{z}_i = {\varphi(\mathbf{x}_i)}/{\| \varphi(\mathbf{x}_i)\|}$ for $i = 1,\dots, n$
      \FOR{$\mathbf{z}_i, i = 1,\dots, n$}
          \STATE compute the distance to $k$-th nearest neighbor of $\mathbf{z}_i$,  $d_i = \| \mathbf{z}_i - \mathbf{z}^{(k)} \| $
      \ENDFOR
      \STATE Sort the distances $\mathbf{D} = \{d^{(k)}_i, i = 1,\dots, n\}$ in ascending order.
      \STATE Set threshold $T$ to the $(1 - \alpha)*n$ order statistic of $D$:
      \STATE $\quad\quad\quad T = d^{(K)}\quad where \quad K = \lfloor(1 - \alpha)*n\rfloor$
   \end{algorithmic}
   \begin{algorithmic}
   \STATE \underline{Test point evaluation:}
   \STATE Given test point $\mathbf{x}_T$, user specified $k$, encoder $\varphi$, threshold $T$, and  reference latent space points $\mathbf{Z} = [ \mathbf{z}_1, \mathbf{z}_2, \dots, \mathbf{z}_n] $
   \STATE Compute $ \mathbf{z}_T = {\varphi(\mathbf{x}_T)}/{\| \varphi(\mathbf{x}_T)\|}$
   \STATE Compute the distance to $k$-th nearest neighbor in $\mathbf{Z}$ of $\mathbf{z}_T$,  $d_T = \| \mathbf{z}_T - \mathbf{z}^{(k)} \| $
   \IF{$d_T \le T$}
   \STATE $\mathbf{x}_T$ is {\it{in-distribution}}
   \ELSE 
   \STATE $\mathbf{x}_T$ is {\it{out-of-distribution}}
   \ENDIF
   \end{algorithmic}
\end{algorithm}

\section{Experiment Details}
\label{app:experiments} 
In this Appendix we provide additional detail about the processing environment used for the training described in Section~\ref{sec:experiments}.

The specific python environment and major packages leveraged for the experiments are as follows:
\begin{itemize}
  \item Python 3.12
  \item Tensorflow 2.18
  \item Keras 3.8
  \item numpy 2.02
  \item matplotlib 3.10
  \item seaborn 0.13
\end{itemize}

Processing requirements for the experiments were not extreme. The workstation used to training the models has the following characteristics:
\begin{itemize}
  \item Intel Xeon Gold 6258BR CPU operating at 2.7 GHz and having 1.5TB of RAM.
  \item Nvidia RTX 6000 GPU with 48GB of RAM.
\end{itemize}
Training of each model instance took less that 1 hour of wall clock time.

\section{Representative Model Selection}
\label{app:modelselection}

Figure~\ref{fig:SarAccuracyFig} presents the classification performance of each of the 50 augmentation-trained ResNet instances on the MSTAR subset ($16.5^\circ - 17.5^\circ$ el, $10 - 80^\circ$ az).  Note that is the validation set used for augmentation parameter optimization.  Table~\ref{tab:AccuracySummaryTable} lists summary statistics for the SAR datasets across all 50 models.

 \begin{figure} [ht]
   \begin{center} 
    \includegraphics[width=5.0in]{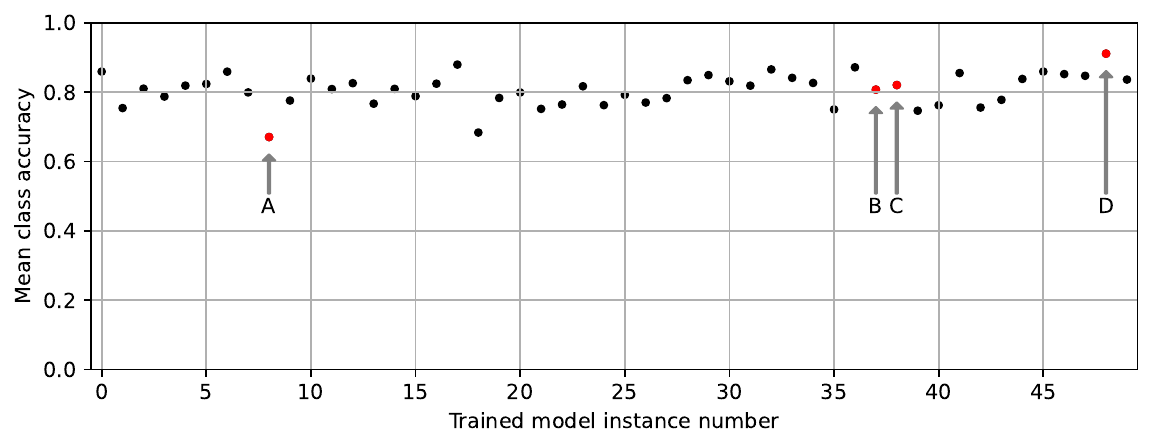}
   \end{center}
   \caption{\label{fig:SarAccuracyFig} MSTAR validation set accuracies for the ResNet-20 models trained with additive Gaussian noise augmentation ($\sigma_{noise} = 1.2$) }
 \end{figure}

\begin{table}[h!]
\caption{Summary statistics for the SAR dataset accuracies across all 50 augmentation-trained ResNet-20 model instances  ($\sigma_{noise} = 1.2$)} 
\label{tab:AccuracySummaryTable}
\begin{center} 
\footnotesize
\begin{tabular}{p{2in}  |  c  | c  |  c  | c |}
\cline{2-5}
   &                &   \multicolumn{3}{c|}{ Mean class accuracy (50 instances)} \\
   &  Set size  & Minimum  &  \hspace{0.2cm} Mean   \hspace{0.2cm} &  Maximum  \\
\cline{1-5}
Augmented SAMPLE  & 10,000  & \multirow{2}{*} {0.901} & \multirow{2}{*}{0.962} & \multirow{2}{*}{0.984} \\
($13.5 - 16.5^\circ$ el, $10 - 80^\circ$ az)  & images &  &  & \\
\cline{1-5}
Augmented SAMPLE  & 10,000  & \multirow{2}{*}{0.877} & \multirow{2}{*}{0.941} & \multirow{2}{*}{0.970}  \\
($16.5 - 17.5^\circ$ el, $10 - 80^\circ$ az)  & images &  &  & \\
\cline{1-5}
SAMPLE ($13.5 - 17.5^\circ$ el, & 1,345 & \multirow{2}{*}{0.729} & \multirow{2}{*}{0.944} & \multirow{2}{*}{0.996}  \\
$10 - 80^\circ$ az)  & images & & &     \\
\cline{1-5}
MSTAR ($13.5 - 17.5^\circ$ el, & 1,345 & \multirow{2}{*}{0.699} & \multirow{2}{*}{0.789 } & \multirow{2}{*}{0.865}  \\
$10 - 80^\circ$ az)  &  images & & &     \\
\cline{1-5}
MSTAR ($29.5 - 30.5^\circ$ el, $10 - 80^\circ$ az,& 114 &  \multirow{2}{*}{0.105} & \multirow{2}{*}{0.539} & \multirow{2}{*}{0.956}  \\
 2 common classes)   & images & & &     \\
 \cline{1-5}
\end{tabular}
\end{center}
\end{table} 

To simplify our analysis, we selected four representative ResNet model instances from that span the accuracies on the MSTAR validation set.  These four instances will be denoted as model instances A through D, where
\begin{itemize}  \item  Instance D (48) is the model instance with the highest validation set accuracy (0.91),
  \item  Instances C and B (38 and 37) have intermediate level validation set accuracies  (0.82 and 0.81, respectively),
  \item  Instance A (8) with the lowest validation accuracy (0.67).
\end{itemize}
While some analysis of classification performance is discussed on all 50 models, the in-depth analysis of outlier detection and classification performance is focused on these four models.

\section{P-Value Analysis}
\label{app:pvalue}

An additional view into the correlation between the percentage \enquote{in-distribution} latent space data and classification accuracy across all fifty instances are given in Figure~\ref{fig:AccVsInliers}.  Additionally, a formal calculation of Pearson correlation with 2-sided $p$-values between these measures are calculated for each dataset across the fifty instances.  The resulting Pearson correlation coefficients are tabulated in Table~\ref{tab:PearsonR}.  Only the ($13.5 - 17.5^\circ$ el) MSTAR estimates a statistically significant correlation between the two variables for the fifty models.  The rest of the datasets fail to demonstrate any meaningful linear correlation.

\begin{figure} [h!]
   \begin{center}
   \begin{tabular}{c} 
    \includegraphics[width=5.0in]{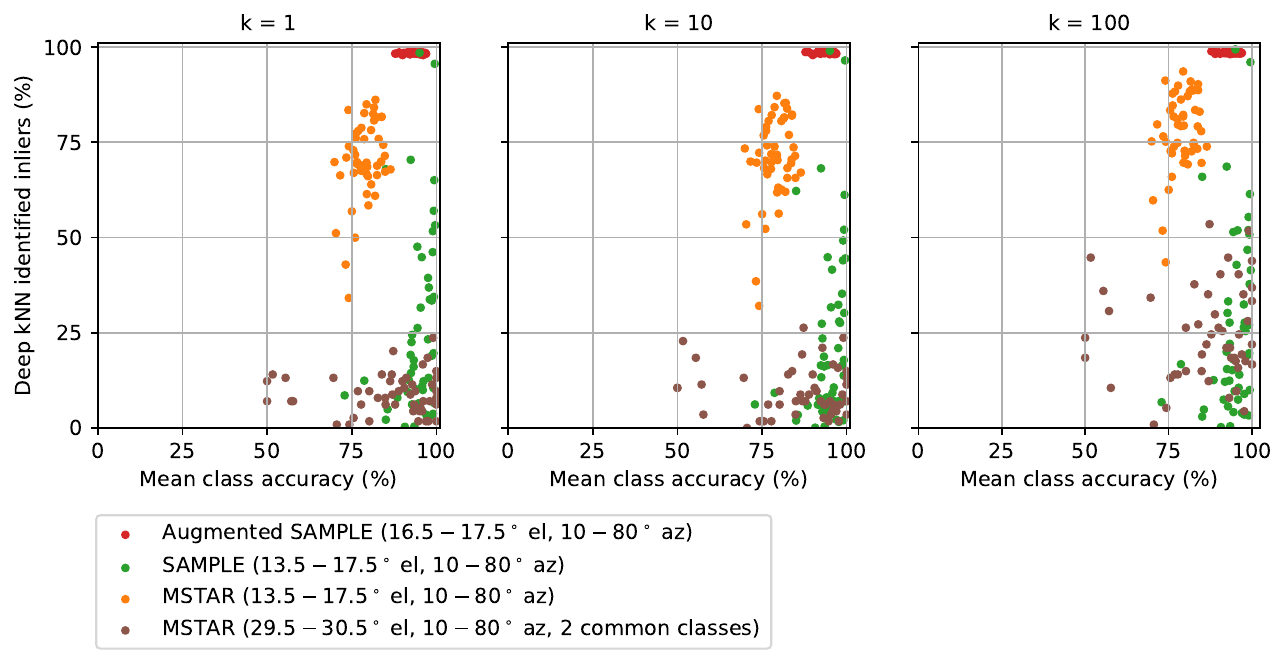}
   \end{tabular}
   \end{center}
   \caption{\label{fig:AccVsInliers} Scatter plots of model instance mean class accuracies versus deep kNN identified inliers for the 4 SAR datasets, across all 50 model instances  ($k = \{1, 10, 100\}$  and $\alpha = 0.1$)  }
    \end{figure} 

 \begin{table}[h!]
\caption{Pearson $r$ between mean class accuracy and deep kNN inlier percentage for 50 model instances ($k = 1, 10, 100; \,\,\alpha = 0.01$)} 
\label{tab:PearsonR}
\begin{center}
\footnotesize
\begin{tabular}{p{1.7in}  |c|c|c|c|c|c|}
\cline{2-7}
   &  \multicolumn{2}{c|}{$k=1$ }  & \multicolumn{2}{c|}{$k=10$ } & \multicolumn{2}{c|}{$k=100$ }  \\                                         
   &  $r$    &    2-sided $p$  &  $r$    &  2-sided $p$  &  $r$    &   2-sided $p$    \\
\cline{1-7}
Augmented SAMPLE                                                                              & \multirow{2}{*}{-0.13} & \multirow{2}{*}{0.36} & \multirow{2}{*}{-0.22} & \multirow{2}{*}{0.12} & \multirow{2}{*}{-0.16} & \multirow{2}{*}{0.27} \\
($16.5 - 17.5^\circ$ el, $10 - 80^\circ$ az)                                              &          &         &          &         &         &         \\
\cline{1-7}
SAMPLE  ($13.5 - 17.5^\circ$ el,                                                            & \multirow{2}{*}{0.26} & \multirow{2}{*}{0.07} &  \multirow{2}{*}{0.26} & \multirow{2}{*}{0.07} &  \multirow{2}{*}{0.26} & \multirow{2}{*}{0.07} \\
$10 - 80^\circ$ az)                                                                                   &          &         &          &         &         &         \\
\cline{1-7}
MSTAR  ($13.5 - 17.5^\circ$ el,                                                              & \multirow{2}{*}{0.36} & \multirow{2}{*}{0.01} &  \multirow{2}{*}{0.30} & \multirow{2}{*}{0.03} &  \multirow{2}{*}{0.33} & \multirow{2}{*}{0.02} \\
$10 - 80^\circ$ az)                                                                                  &          &         &          &         &         &         \\
\cline{1-7}
MSTAR  ($29.5 - 30.5^\circ$ el,                                                              & \multirow{2}{*}{0.02} & \multirow{2}{*}{0.90} & \multirow{2}{*}{-0.08} & \multirow{2}{*}{0.57} &  \multirow{2}{*}{0.06} & \multirow{2}{*}{0.70} \\
$10 - 80^\circ$ az, 2 common classes)                                                   &          &         &          &         &         &         \\
\cline{1-7}
\end{tabular}
\end{center}
\end{table} 

\end{appendices}

\end{document}